\documentclass[10pt, a4paper]{article}

\usepackage[final]{lrec-coling2024} 

\usepackage{natbib}
\usepackage{multibib}
\makeatletter
\def\@mb@citenamelist{cite,citep,citet,citealp,citealt,citepalias,citetalias}
\makeatother
\newcites{languageresource}{~}

\usepackage{mathtools}
\usepackage{graphicx}
\usepackage{tabularx}
\usepackage{soul}
\usepackage{multirow}
\usepackage{microtype}
\usepackage{latexsym}
\usepackage{amsmath}
\usepackage{url}
\usepackage{amssymb}
\usepackage{amsfonts}
\usepackage{graphicx}
\usepackage{tabularx}
\usepackage{multirow}
\usepackage{arydshln}
\usepackage{mathtools,nccmath}
\usepackage{enumitem}
\usepackage{todonotes}
\usepackage{listings}
\usepackage{xcolor}
\usepackage{hyperref}
 \definecolor{darkblue}{rgb}{0, 0, 0.5}
\hypersetup{colorlinks=true, citecolor=darkblue, linkcolor=darkblue, urlcolor=darkblue}

\usepackage{xstring}

\usepackage{color}

\newcommand{\ignore}[1]{}
\title{Improving Vietnamese-English Medical Machine Translation}

\name{Nhu Vo$^1$, Dat Quoc Nguyen$^2$, Dung D. Le$^1$, Massimo Piccardi$^3$, Wray Buntine$^1$} 

\address{$^1$College of Engineering and Computer Science, VinUniversity, Vietnam;\\
$^2$VinAI Research, Vietnam; $^3$University of Technology Sydney, Australia\\
         $^1$\{nhu.vd, dung.ld, wray.b\}@vinuni.edu.vn; $^2$v.datnq9@vinai.io; $^3$massimo.piccardi@uts.edu.au
         \\}

\abstract{
Machine translation for Vietnamese-English in the medical domain is still an under-explored research area. In this paper, we introduce MedEV---a  high-quality Vietnamese-English parallel dataset constructed specifically for the medical domain, comprising approximately 360K sentence pairs. We conduct extensive experiments comparing Google Translate, ChatGPT (gpt-3.5-turbo), state-of-the-art Vietnamese-English neural  machine translation models and pre-trained bilingual/multilingual sequence-to-sequence models on our new MedEV dataset. Experimental results show that the best performance is achieved by fine-tuning \texttt{vinai-translate} \cite{vinaitranslate} for each translation direction. We publicly release our dataset to promote further research.
 \\ \newline 
}

\begin{document}

\maketitleabstract

\section{Introduction}

Almost all medical universities and hospitals in Vietnam use Vietnamese in their teaching and practices. Additionally, the majority of specialized educational materials created for students, doctors, and nurses are in English. Even though some undergraduate and many higher-degree medical programs now incorporate English, learners are still required to use Vietnamese in their daily professional interactions.  Thus, the demand for high-quality Vietnamese-English medical machine translation (MT) has increased significantly.

For training an MT model, a suitable parallel dataset is needed \cite{el-kishky-etal-2020-ccaligned, schwenk-etal-2021-wikimatrix, nguyen-etal-2022-kc4mt}. Previous Vietnamese-English data comes from publicly available resources \cite{TIEDEMANN12.463,cettolo-etal-2015-iwslt}, and then a particular methodology for creating parallel sentences is followed \cite{PhoMT,PhoST}.   There are two prominent high-quality and large-scale Vietnamese-English parallel datasets that have been made publicly available to date: PhoMT \cite{PhoMT} and MTet \cite{Ngo2022MTetMT}. However, PhoMT does not contain pairs from the medical domain, while MTet contains 13,410 medical sentence pairs. In addition, the COVID-19 - HEALTH Wikipedia dataset contains 4,273 Vietnamese-English sentence pairs in COVID-19 news.\footnote{\url{https://www.elrc-share.eu}} These numbers of medical sentence pairs are small for high-quality medical translation training. This is a compelling motivation for the development of a dedicated high-quality Vietnamese-English parallel dataset to bridge the gap in the available resources for machine translation in the medical domain.


In this paper, as our first contribution, we introduce the MedEV dataset, a high-quality Vietnamese-English parallel corpus containing 358.7K sentence pairs in the medical domain. As our second contribution, we conduct a comprehensive empirical investigation using the MedEV dataset to improve the performance of neural machine translation (NMT) models within the medical health domain. In particular, we compare the performance of medical text translation among various translation tools and models, including Google Translate, ChatGPT (gpt-3.5-turbo), state-of-the-art Vietnamese-English NMT models, and pre-trained bilingual/multilingual sequence-to-sequence models. To the best of our knowledge, this marks the first empirical study focusing on Vietnamese-English medical machine translation.

We make the MedEV dataset publicly available for research and educational purposes.\footnote{Our  MedEV dataset is publicly available at: \url{https://huggingface.co/datasets/nhuvo/MedEV}.} We hope that MedEV, along with our empirical study, will serve as a foundational resource for future research and applications in the field of Vietnamese-English medical machine translation.





\section{Our MedEV Dataset}

\begin{table*}[!t]
\centering
\resizebox{16cm}{!}{
\setlength{\tabcolsep}{0.33em}
\def\arraystretch{1.1}
\begin{tabular}{l|l|l|l|l|l|l|l|l|l|l|l|l|l|l}
\hline
\multirow{2}{*}{\textbf{Genre}} & \multicolumn{2}{c|}{\textbf{Total}}& \multicolumn{4}{c|}{\textbf{Training}}& \multicolumn{4}{c|}{\textbf{Validation}}& \multicolumn{4}{c}{\textbf{Test}} \\
\cline{2-15}
&{\textbf{\#doc}}& {\textbf{\#pair}} & {\textbf{\#doc}} &
 {\textbf{\#pair}} & {\textbf{\#en/s}} & {\textbf{\#vi/s}} & {\textbf{\#doc}} & {\textbf{\#pair}} & {\textbf{\#en/s}} & {\textbf{\#vi/s}} & {\textbf{\#doc}} & {\textbf{\#pair}} & {\textbf{\#en/s}} & {\textbf{\#vi/s}}\\
\hline
Article Abstracts & 22,580 & 196,276 & 21,397 & 186,528 & 23.87& 31.67& 583 & 4,883 & 24.38 & 32.32 & 600 & 4,865 &24.51 &32.24\\
MSD Manuals & 2,796 & 123,302 & 2,652& 117,101 & 25.96& 36.39& 87& 3,079 & 26.98 & 37.98 & 57& 3,122 &26.67 &37.30\\
Thesis Summaries & 783 & 23,084& 731 & 21,940& 30.27& 36.9 & 35& 571& 28.83 & 37.33 & 17& 573&25.40 &31.92\\
Article Translations & 1,059& 16,134& 1,000& 15,328& 25.92& 34.56& 31& 406& 25.57 & 33.83 & 28& 400&26.30 &35.41\\
\hline
All & 27,218 & 358,796 & 25,780 & 340,897 & 25.09& 33.76& 736 & 8,939 & 25.61& 34.65 & 702 & 8,960 & 25.4 & 34.12 \\
\hline
\end{tabular}
}
\caption{Statistics of our MedEV dataset. \textbf{\#doc}: The number of parallel document pairs. \textbf{\#pair}: The number of parallel sentence pairs. \textbf{\#en/s}: The average number of word tokens per English sentence. \textbf{\#vi/s}: The average number of syllable tokens per Vietnamese sentence.}
\label{tab:stats}
\end{table*}

Developing our MedEV dataset involves three main stages. First, we collect parallel document pairs in the medical domain and then preprocess the collected data. Second, we perform the alignment of parallel sentences within pairs of parallel documents. Last, we perform post-processing steps, which include removing duplicate sentences and manually verifying the quality of the validation and test splits.


\subsection{Data collection and pre-processing}

We collect 27,218 parallel document pairs from publicly accessible resources across four genres, including: (i) 22,580 bilingual Vietnamese-English abstracts derived from scientific articles published in medical, clinical, and pharmaceutical journals based in Vietnam; (ii) 2,796 English documents and their corresponding Vietnamese-translated versions from the MSD Manuals website;\footnote{\url{https://www.msdmanuals.com/professional}} (iii) 783 bilingual Vietnamese-English summaries extracted from doctoral dissertations from official websites of medical universities in Vietnam; and (iv) 1,059 English scientific articles and their Vietnamese translations, completed by Vietnamese medical doctors. 

Here, these document pairs are available in either HTML web pages or in PDF/DOC/DOCX files. To process HTML web pages, we crawl and extract parallel text pairs using the DownThemAll\footnote{\url{https://www.downthemall.org/}} tool and the ``BeautifulSoup'' library.\footnote{\url{https://pypi.org/project/beautifulsoup4/}} For PDF/DOC/DOCX files, we download and convert them into the plain text format.\footnote{We use the ``pdftotext'' Python library to extract content from PDF files, typically formatted in two columns.}  Afterward, we manually eliminate unnecessary elements such as headers, footers, footnotes, and page numbers from articles, and then extract the bilingual abstract/summary pairs.



To extract sentences for the next stage of parallel sentence alignment, we automatically segment each text document into sentences, using the Stanford CoreNLP toolkit for English \cite{manning-etal-2014-stanford} and the VnCoreNLP toolkit for Vietnamese \cite{nguyen-etal-2018-fast,vu-etal-2018-vncorenlp}.


\subsection{Sentence pair alignment}

Following the PhoMT alignment approach \cite{PhoMT}, we align parallel sentences within a parallel document pair, as follows:  (\textbf{1}) Translate each English source sentence into Vietnamese by using the pre-trained model \texttt{vinai-translate}  \cite{vinaitranslate}.\footnote{\url{https://github.com/VinAIResearch/VinAI_Translate}}
(\textbf{2}) Align English-Vietnamese sentence pairs via an ``intermediate'' alignment between the Vietnamese-translated versions of the English source sentences and the Vietnamese target sentences. This is done by using alignment toolkits Hunalign \cite{Varga2007ParallelCF} and Bleualign \cite{Sennrich2011IterativeMS}.
(\textbf{3}) Select sentence pairs that were aligned by both of these alignment toolkits.

Hunalign and Bleualign include 99\% and 95\% of Vietnamese/English sentences from our raw dataset into their output, respectively, resulting in an alignment coverage rate of 93+\% of Vietnamese/English sentences to be included in the alignment output of about 390K sentence pairs.


\subsection{Data post-processing}

Out of the 390K English-Vietnamese sentence pairs generated in the previous stage, we exclude 14K sentence pairs with SacreBLEU scores \cite{Post2018ACF} falling outside the range of $[5, 95)$. Subsequently, we also remove 16K duplicate sentence pairs, both within and across all document pairs, resulting in a dataset of 358,885 unique sentence pairs.
This dataset is randomly split at the document level, following a sentence pair ratio of 0.95 / 0.025 / 0.025, thus yielding a total of 340,897 sentence pairs for training, 8,982 for validation, and 9,006 for test.


To assess the dataset's quality, we conduct a manual examination within our validation and test sets. This evaluation task is carried out by two third-year medical undergraduates,\footnote{Examiners have a proficient English level at IELTS 7.0+ and GPA 3.5+/4.0.} who are responsible for determining if each sentence pair is misaligned (i.e. completely different sentence meaning or partly preserving the sentence meaning). Each examiner independently assesses a total of 8,982 + 9,006 = 17,988 sentence pairs within an average of 90 hours. Then, we perform a cross-checking process and find that 43 validation sentence pairs (0.48\%) and 46 test sentence pairs (0.51\%) exhibits misalignment. Given the tiny percentage of misalignment at the sentence level in both the validation and test sets, we assert that our training set maintains a high standard of quality. Finally, we remove those misaligned pairs, resulting in a final count of 8,939 high-quality sentence pairs for validation and 8,960 for test. Table \ref{tab:stats} shows the statistics of our MedEV dataset.

\section{Experiment Setup}
\subsection{Experimental models}

Our experimental setup focuses on using the MedEV dataset to explore: (i) the dataset's quality as demonstrated by its usage in improving neural machine translation (NMT) models' performance in the medical health domain; and (ii) a comparison of medical text translation performance among a well-known translation engine - Google Translate, a large language model - ChatGPT (gpt-3.5-turbo), pre-trained multilingual translation models \texttt{SeamlessM4T} \cite{communication2023seamlessm4t} and \texttt{M2M100} \cite{10.5555/3546258.3546365}, state-of-the-art Vietnamese-English NMT models \texttt{vinai-translate} \cite{vinaitranslate} and \texttt{envit5-translation} \cite{Ngo2022MTetMT}, 
and pre-trained sequence-to-sequence models mBART  \cite{Liu2020MultilingualDP} and \texttt{envit5-base} \cite{Ngo2022MTetMT}.

mBART is pre-trained on a dataset of 25 languages, that contains 300GB of English texts and 137 GB of Vietnamese texts. Subsequently,  \texttt{vinai-translate} is  fine-tuned using mBART on a dataset of 9M sentence pairs, including 3M high-quality pairs in PhoMT \cite{PhoMT} and an additional 6 million pairs from the noisier datasets CCAligned \cite{el-kishky-etal-2020-ccaligned} and WikiMatrix \cite{schwenk-etal-2021-wikimatrix}. On the other hand, \texttt{envit5-base} is a bilingual variant of the T5  model \cite{t5}, pre-trained on a dataset consisting of 80GB of English texts and 80GB of Vietnamese texts. Furthermore,  \texttt{envit5-translation} is  fine-tuned using \texttt{envit5-base} on a dataset of 6.2M high-quality sentence pairs from both PhoMT and the MTet dataset \cite{Ngo2022MTetMT}. 


\subsection{Implementation details}

On our MedEV dataset, we fine-tune the models \texttt{vinai-translate}, \texttt{envit5-translation}, mBART, and \texttt{envit5-base} for 5 epochs with  AdamW  \cite{loshchilov2017decoupled}, using HuggingFace ``transformers'' library \cite{wolf-etal-2020-transformers}. We use an initial learning rate of 5e-5 and a maximum sequence length of 256. We employ mixed precision training (fp16), using 4 NVIDIA A100 GPUs, a batch size of 4 for each GPU, with 8 steps of gradient accumulation and 1250 warm-up steps.


We use beam search with a beam size of 5 for decoding. The performance is computed using metrics  BLEU \cite{Papineni2002BleuAM}, TER \cite{snover-etal-2006-study} and METEOR \cite{banarjee2005}. Here, we calculate the case-sensitive BLEU score using SacreBLEU \cite{Post2018ACF}. 
Each model is evaluated after every 1000 training steps, and the model checkpoint that yields the highest BLEU score on the validation set is selected for evaluation on the test set.

For ChatGPT (gpt-3.5-turbo), we conduct zero-shot/few-shot ``in-context'' learning. In the $n$-shot setting, we randomly select $n$ samples from the training set for the prompt content for each validation/test sample. Note that for $n=32$, since the ``gpt-3.5-turbo'' model limits requests to 4096 tokens, we restrict randomly sampled training sentences with a length of fewer than 64 tokens. Please refer to the prompt construction template in the Appendix \ref{appendix}. 
In a preliminary experiment, we find that a temperature value of 0.2 yields the best performance score. Therefore, we report all our ChatGPT results using a fixed temperature of 0.2. 



\begin{table}[!t]
\centering
\resizebox{7.5cm}{!}{
\setlength{\tabcolsep}{0.25em}
\def\arraystretch{1.1}
\begin{tabular}{ll|c|c|c|c}
\hline
& \multirow{2}{*}{\textbf{Model}} & \multicolumn{2}{c|}{\textbf{Validation set}}& \multicolumn{2}{c}{\textbf{Test set}} \\
\cline{3-6}
\cline{3-6}
& & \textbf{En2Vi}& \textbf{Vi2En}& \textbf{En2Vi}& \textbf{Vi2En}\\
\hline
\multirow{8}{*}{\rotatebox[origin=c]{90}{\textbf{w/o FT}}}
& Google Translate & {47.37} & {38.50} & {47.86} & {39.26} \\
& ChatGPT 0-shot & 34.38 & 29.79 & 34.45 & 30.39 \\
& ChatGPT 1-shot & 35.28 & 31.27 & 35.23 & 31.70 \\
& ChatGPT 8-shot & 36.09 & 31.87 & 36.02 & 32.57 \\
& ChatGPT 16-shot & 36.32 & 32.14 & 35.69 & 32.90 \\
& ChatGPT 32-shot & 34.92 & 32.08 & 36.37 & 32.94 \\
& SeamlessM4T medium & {31.04} & {21.57} & {31.25} & {21.65} \\
& M2M100 418M & {28.30} & {22.46} & {28.26} & {22.56} \\
\cline{2-6}
& vinai-translate &44.24& 33.28& 44.60&33.44\\
& envit5-translation &42.86& 31.33& 43.23& 32.00\\
\hline
\hline
\multirow{4}{*}{\rotatebox[origin=c]{90}{\textbf{FT}}} 
& vinai-translate & \textbf{52.21} & \textbf{42.66} & \textbf{52.14} & \textbf{42.38} \\
& envit5-translation & 51.14 & 41.47 & 51.27 & 41.17 \\
& mBART & 51.23 & 41.67 & 51.18 & 41.51\\
& envit5-base & 50.10 & 40.66 & 49.94 & 40.36\\
\hline
\end{tabular}
}
\caption{BLEU scores. ``FT'' denotes fine-tuning.}
\label{tab:results}
\end{table}

\setcounter{table}{3}
\begin{table*}[!t]
\centering
\resizebox{15cm}{!}{
\setlength{\tabcolsep}{0.25em}
\def\arraystretch{1.1}
\begin{tabular}{ll|c|c|c|c|c|c|c|c|c|c|c|c}
\hline
&\multirow{3}{*}{\textbf{Model}} & \multicolumn{6}{c|}{\textbf{English-to-Vietnamese}} & \multicolumn{6}{c}{\textbf{Vietnamese-to-English}} \\
\cline{3-14}
&& {\textbf{$<$ 10}} & {\textbf{[10, 20)}} & {\textbf{[20, 30)}} & {\textbf{[30, 40)}} & {\textbf{[40, 50)}} & {\textbf{$\geq$ 50}} & {\textbf{$<$ 10}} & {\textbf{[10, 20)}} & {\textbf{[20, 30)}} & {\textbf{[30, 40)}} & {\textbf{[40, 50)}} & {\textbf{$\geq$ 50}} \\
&& 5.16\% & 24.96\% & 28.40\% & 19.59\% & 10.40\% & 11.48\%& 13.99\% & 39.85\% & 26.94\% & 11.27\% & 3.93\% & 4.02\% \\
\hline
\multirow{8}{*}{\rotatebox[origin=c]{90}{\textbf{w/o FT}}}
&Google Translate &43.17 & 45.16 & 47.33 & 48.20 & 48.57 & 48.16 & 34.08 & 37.86 & 38.97 & 39.47 & 41.01 & 42.41  \\ 

&ChatGPT 0-shot & 30.46 & 32.15 & 33.39 & 34.67 & 35.55 & 35.61& 26.15 & 27.53 & 29.73 & 31.63 & 34.07 & 35.54 \\
&ChatGPT 1-shot & 31.67 & 33.22 & 34.16 & 35.49 & 36.17 & 35.87& 27.32 & 28.78 & 31.20 & 33.05 & 34.99 & 36.75 \\
&ChatGPT 8-shot & 34.08 & 34.06 & 35.19 & 36.23 & 36.69 & 36.28& 27.97 & 29.72 & 32.15 & 33.38 & 35.85 & 37.75 \\
&ChatGPT 16-shot & 29.95 & 32.91 & 34.93 & 36.02 & 36.60 & 36.34& 28.10 & 29.97 & 32.42 & 33.89 & 35.93 & 38.34 \\
&ChatGPT 32-shot & 34.94 & 34.82 & 35.42 & 36.67 & 37.18 & 36.18& 28.39 & 30.04 & 32.49 & 33.87 & 36.27 & 38.11 \\
& SeamlessM4T medium & 25.78 & 29.30 & 30.58 & 32.20 & 32.60 & 29.81 & 16.06 & 19.82 & 22.35 & 22.93 & 24.54 & 19.74 \\
& M2M100 418M & 24.07 & 27.08 & 28.07 & 29.04 & 29.66 & 27.14 & 19.40 & 20.55 & 22.56 & 23.98 & 24.59 & 24.17\\
\cline{2-14}
&vinai-translate & 31.53 & 43.07 & 44.51 & 44.77 & 43.70 & 43.92& 28.81 & 30.99 & 33.03 & 34.36 & 36.01 & 38.01 \\
&envit5-translation & 38.72 & 41.77 & 42.75 & 43.73 & 44.08 & 42.59& 27.07 & 28.31 & 31.76 & 33.89 & 35.53 & 37.12 \\
\hline
\hline
\multirow{4}{*}{\rotatebox[origin=c]{90}{\textbf{FT}}}
&vinai-translate & 48.64 & \textbf{50.58} & \textbf{50.93} & \textbf{51.59} & \textbf{51.63} & \textbf{52.92}& \textbf{38.07} & \textbf{39.97} & \textbf{41.24} & \textbf{41.80} & \textbf{44.59} & 47.12  \\
&envit5-translation & \textbf{49.97} & 50.50 & 50.30 & 50.81 & 51.27 & 51.99& 35.32 & 38.07 & 40.11 & 41.32 & 44.44 & \textbf{47.28} \\
&mBART & 48.85 & 49.83 & 50.18 & 50.43 & 51.00 & 51.61& 37.88 & 38.91 & 40.44 & 40.22 & 43.89 & 46.01 \\
&envit5-base & 49.11 & 49.13 & 48.95 & 48.88 & 49.12 & 49.98 & 35.43 & 37.62 & 39.05 & 38.89 & 42.02 & 44.14 \\
\hline 
\end{tabular}
}
\caption{BLEU scores on the test set w.r.t. sentence lengths of reference sentences (i.e. the number of words including punctuations). The number below each length bucket indicates the percentage of sentences in that bucket.}
\label{tab:bleu_sent_len_vien}
\end{table*}

\begin{table*}[!t]
\centering
\resizebox{15cm}{!}{
\setlength{\tabcolsep}{0.25em}
\def\arraystretch{1.1}
\begin{tabular}{ll|c|c|c|c|c|c|c|c}
\hline
&\multirow{3}{*}{\textbf{Model}} & \multicolumn{4}{c|}{\textbf{English-to-Vietnamese}} & \multicolumn{4}{c}{\textbf{Vietnamese-to-English}}\\
\cline{3-10}
&& \textbf{Article} & \textbf{MSD} & \textbf{Thesis} & \textbf{Article}& \textbf{Article} & \textbf{MSD} & \textbf{Thesis} & \textbf{Article}\\
&& \textbf{Abstracts} & \textbf{Manuals} & \textbf{Summaries} & \textbf{Translations}& \textbf{Abstracts} & \textbf{Manuals} & \textbf{Summaries} & \textbf{Translations}\\
\hline
\multirow{8}{*}{\rotatebox[origin=c]{90}{\textbf{w/o FT}}}
&Google Translate & 40.06 & 56.86 & 49.20 & 52.79& 32.17 & 48.38 & \textbf{44.82} & 40.92 \\ 
&ChatGPT 0-shot & 30.48 & 39.05 & 34.18 & 39.59& 25.79 & 36.14 & 31.94 & 35.14 \\
&ChatGPT 1-shot & 31.42 & 39.66 & 35.02 & 39.78& 26.69 & 38.14 & 33.25 & 35.75 \\
&ChatGPT 8-shot & 32.40 & 40.23 & 36.08 & 40.13& 27.26 & 39.46 & 34.08 & 36.23 \\
&ChatGPT 16-shot & 31.91 & 39.96 & 36.28 & 40.38& 27.50 & 40.08 & 33.85 & 36.13 \\
&ChatGPT 32-shot & 32.97 & 40.30 & 36.77 & 40.51& 27.37 & 40.23 & 34.41 & 36.52 \\
& SeamlessM4T medium & 25.56 & 38.02 & 28.01 & 40.32 & 17.94 & 26.09 & 21.63 & 28.92  \\
& M2M100 418M & 23.13 & 34.36 & 24.35 & 37.69 & 19.36 & 26.20 & 23.77 & 28.30  \\
\cline{2-10}
&vinai-translate & 37.99 & 53.46 & 37.03 & 48.74& 28.07 & 39.79 & 35.82 & 39.34  \\
&envit5-translation & 37.44 & 50.85 & 41.04 & 46.89& 24.51 & 42.86 & 33.65 & 38.13 \\
\hline
\hline
\multirow{4}{*}{\rotatebox[origin=c]{90}{\textbf{FT}}}
&vinai-translate & \textbf{45.69} & \textbf{60.77} & \textbf{50.74} & \textbf{50.92}& \textbf{33.25} & \textbf{54.54} & 42.22 & \textbf{41.86} \\
&envit5-translation & 44.73 & 60.29 & 50.02 & 50.09& 32.32 & 54.26 & 40.54 & 37.92  \\
&mBART & 45.54 & 59.18 & 50.21 & 45.54& 33.13 & 52.83 & 41.86 & 36.54 \\
&envit5-base & 43.58 & 58.13 & 48.16 & 44.00& 32.08 & 51.09 & 39.11 & 36.07 \\
\hline 
\end{tabular}
}
\caption{BLEU scores on the test set for each genre.}
\label{tab:genres_vien}
\end{table*}

\setcounter{table}{2}
\begin{table}[!t]
\centering
\resizebox{7.5cm}{!}{
\setlength{\tabcolsep}{0.25em}
\def\arraystretch{1.1}
\begin{tabular}{ll|c|c|c|c}
\hline
& \multirow{2}{*}{\textbf{Model}} & \multicolumn{2}{c|}{\textbf{En2Vi}}& \multicolumn{2}{c}{\textbf{Vi2En}} \\
\cline{3-6}
& & \textbf{TER\( \downarrow \)}& \textbf{METEOR\( \uparrow \)} & \textbf{TER\( \downarrow \)} & \textbf{METEOR\( \uparrow \)}\\
\hline
\multirow{8}{*}{\rotatebox[origin=c]{90}{\textbf{w/o FT}}}
& Google Translate & 46.30 & 0.704 & 56.52 & 0.665 \\
& ChatGPT 0-shot & 59.35 & 0.625 & 66.68 & 0.608\\
& ChatGPT 1-shot & 58.47 & 0.629 & 64.88 & 0.614\\
& ChatGPT 8-shot & 57.80 & 0.634 & 63.74 & 0.621\\
& ChatGPT 16-shot & 58.57 & 0.629 & 63.46 & 0.622\\
& ChatGPT 32-shot & 57.48 & 0.638 & 63.32 & 0.623\\
& SeamlessM4T medium & {61.69} & {0.576} & {76.13} & {0.498} \\
& M2M100 418M & {64.79} & {0.537} & {75.16} & {0.518} \\
\cline{2-6}
& vinai-translate & 48.69 & 0.685 & 61.93 & 0.626\\
& envit5-translation & 49.98 & 0.673 & 67.63 & 0.627\\
\hline
\hline
\multirow{4}{*}{\rotatebox[origin=c]{90}{\textbf{FT}}} 
& vinai-translate & \textbf{42.22} & \textbf{0.740} &\textbf{52.24} & \textbf{0.685}\\
& envit5-translation & 42.23 & 0.733 & 53.50 & 0.678\\
& mBART & 42.99 & 0.732 & 53.03 & 0.678\\
& envit5-base & 43.43 & 0.720 & 54.07 & 0.666\\
\hline
\end{tabular}
}
\caption{TER and METEOR scores on the test set. }
\label{tab:results_TER_meteor}
\end{table}

\section{Experimental Results}

Tables \ref{tab:results} and \ref{tab:results_TER_meteor} present the BLEU, TER and METEOR scores obtained by all experimental models for both translation directions: English-to-Vietnamese (En2Vi) and Vietnamese-to-English (Vi2En). 
In the  ``without fine-tuning'' (w/o FT) setting, the automatic translation engine Google Translate consistently outperforms both \texttt{vinai-translate} and \texttt{envit5-translation}, achieving the best scores. In contrast, ChatGPT tends to produce lower scores in most cases while \texttt{SeamlessM4T} and \texttt{M2M100 418M} exhibit the poorest performance, significantly behind the superior results of Google Translate. 
This is likely due to Google Translate being trained on some parallel resource in the medical domain. As for ChatGPT, it generally attains better scores when more training pairs are used in the few-shot setups.  
When it comes to the  ``fine-tuning'' setting, all fine-tuned models outperform  Google Translate on both validation and test sets in both translation directions. Here,  \texttt{vinai-translate} achieves the best scores, surpassing Google Translate by a substantial margin. Specifically, it outperforms Google Translate by 4+ BLEU points in English-to-Vietnamese translation and by 3+ BLEU points in Vietnamese-to-English translation.

Tables \ref{tab:bleu_sent_len_vien} and \ref{tab:genres_vien} show BLEU scores on the test set for English-to-Vietnamese and Vietnamese-to-English translation directions regarding each sentence length bucket and resource genre, respectively. We find from Table  \ref{tab:bleu_sent_len_vien} that in medical texts, as the sentence length increases, the probability of encountering common words that match between the machine-translated text and the reference text also increases, resulting in higher BLEU scores. For shorter sentences, the translation system may offer synonymous words or medical terms that do not align perfectly with the reference text. 
As shown in Table \ref{tab:genres_vien}, the highest BLEU scores are reported for MSD Manuals, which are composed of documents written by doctors on common diseases classified under the ICD-10 code system. The following are the scores reported for Thesis Summaries and Article Translations. On the contrary, the remaining resource genre, Article Abstracts (including article titles, abstracts, and keywords), contains more medical terminology than the other categories, resulting in lower BLEU scores. 

\begin{figure}[!t]
\centering
\includegraphics[width=0.95\linewidth]{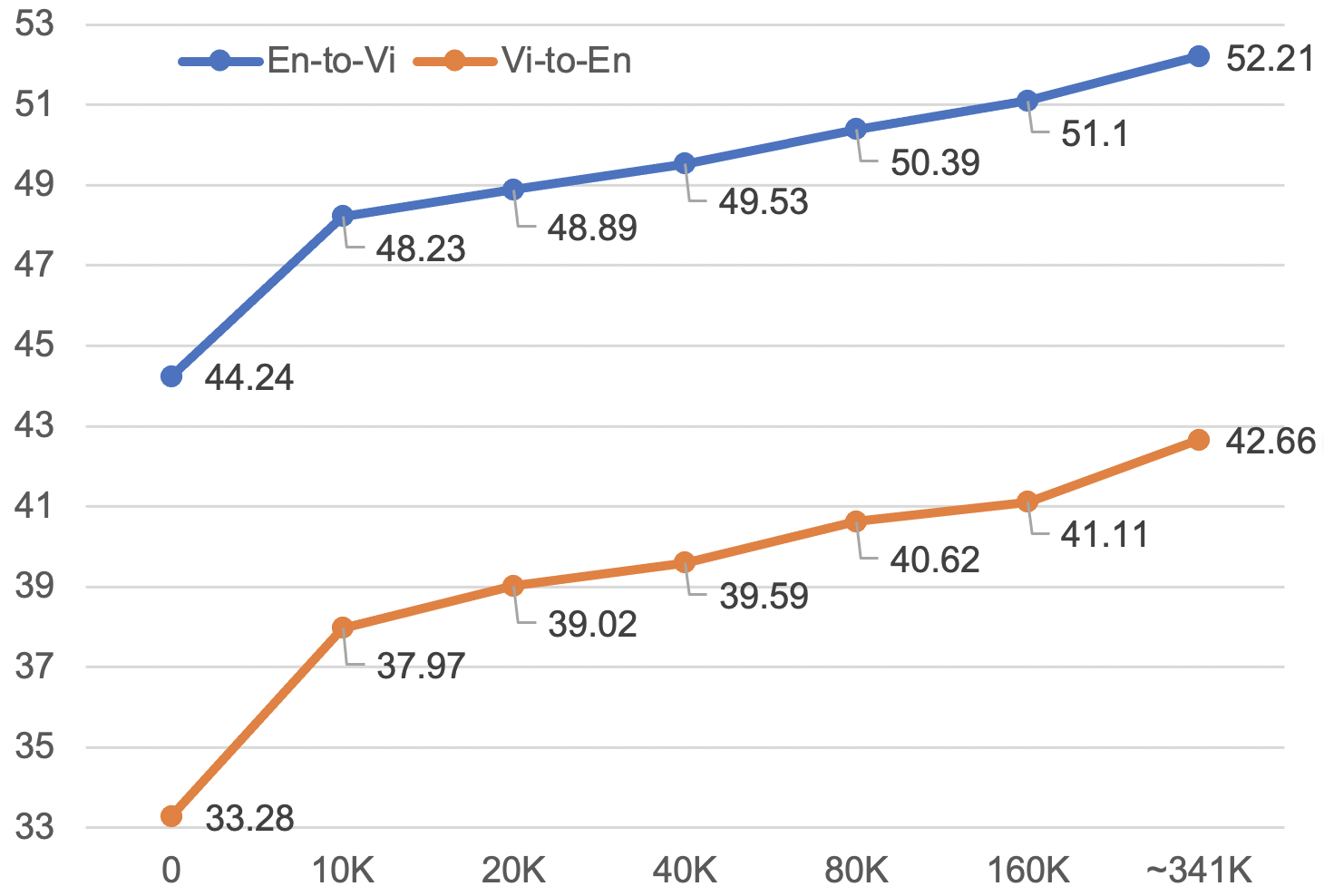}
\caption{BLEU scores on the validation set when fine-tuning \texttt{vinai-translate} with different training sizes for both translation directions.}
\label{fig:model}
\end{figure}

Figure \ref{fig:model} presents BLEU scores on the validation set for both translation directions when fine-tuning \texttt{vinai-translate} with different numbers of training sentence pairs. Here, using only 10K sentence pairs helps substantially improve the baseline scores by 4+ points: from 44.24 to 48.23 for English-to-Vietnamese and from 33.28 to 37.97 for Vietnamese-to-English. Additional 330K+ pairs produce 4+ more points, increasing from 48.23 to 52.21 and from 37.97 to 42.66. These scores clearly demonstrate the positive impacts of larger training sizes.

\section{Conclusion}

In this paper, we have presented a high-quality MedEV dataset of about 360K parallel sentence pairs from 27K documents in the medical domain. We conduct experiments on MedEV to compare strong baselines and demonstrate the effectiveness of the NMT  model \texttt{vinai-translate} in Vietnamese-English medical machine translation. We hope that the public release of our dataset will be a major step in the direction of more extensive Vietnamese-English machine translation in the medical field. In future work, we will explore the translation quality when combining our MedEV with other general domains PhoMT and MTet.


\section{Ethical Statement}


Data are collected from publicly available websites, such as journals and universities, but also from {\tt www.msd.com}.  The content extracted from these sources cannot be used for public or commercial purposes.  Therefore, the content also contains no private data about the patients.

\section{References}\label{sec:reference}
\bibliographystyle{lrec-coling2024-natbib}
\bibliography{refs}

\appendix





\section{Prompt Design}\label{appendix}

\paragraph{Zero-shot Setting:}
\begin{itemize}
    \item For English to Vietnamese translation:
\end{itemize}
\begin{lstlisting} 
I want you to act as a translator to translate text from English to Vietnamese in the medical domain. 
Now with the following English INPUT text:
INPUT= [English input sentence from validation/test set]
What is the corresponding Vietnamese-translated OUTPUT text?
\end{lstlisting}

\begin{itemize}
    \item For Vietnamese to English translation:
\end{itemize}
\begin{lstlisting} 
I want you to act as a translator to translate text from Vietnamese to English in the medical domain. 
Now with the following Vietnamese INPUT text:
INPUT= [Vietnamese input sentence from validation/test set]
What is the corresponding English-translated OUTPUT text?
\end{lstlisting}

\paragraph{Few-shot Setting:}

\begin{itemize}
    \item For English to Vietnamese translation:
\end{itemize}

\begin{lstlisting} 
I want you to act as a translator to translate text from English to Vietnamese in the medical domain. For instance, consider the following English INPUT text:
INPUT= [shot 1 source]
[shot 2 source]
[shot n source]

You would generate a corresponding Vietnamese OUTPUT text as follows:
OUTPUT= [shot 1 reference]
[shot 2 reference]
[shot n reference]

Now with the following English INPUT text:

INPUT= [English input sentence from the validation/test set]

What is the corresponding Vietnamese-translated OUTPUT text?

\end{lstlisting}

\begin{itemize}
    \item For Vietnamese to English translation:
\end{itemize}

\begin{lstlisting} 
I want you to act as a translator to translate text from Vietnamese to English in the medical domain. For instance, consider the following Vietnamese INPUT text:}
INPUT= [shot 1 source]
[shot 2 source]
[shot n source]

You would generate a corresponding English OUTPUT text as follows:
OUTPUT= [shot 1 reference]
[shot 2 reference]
[shot n reference]

Now with the following Vietnamese INPUT text:

INPUT= [Vietnamese input sentence from validation/test set]

What is the corresponding English-translated OUTPUT text?


\end{lstlisting}
The output from the ChatGPT API may sometimes include model-generated sentences in addition to the translation results. We manually check the output and remove these sentences. For instance:
\begin{itemize}
    \item The model repeats sentences from the prompt: \textit{``The corresponding English-translated text is:'', ``The corresponding Vietnamese-translated OUTPUT text is:''}
    \item The model adds new sentences in the response content: \textit{``Possible English translation:'', ``Possible OUTPUT:'', ``Possible translation:''}

\end{itemize}

\end{document}